\documentclass{article}
\usepackage{arxiv}
\usepackage[T1]{fontenc}
\usepackage{hyperref}
\usepackage{booktabs}
\usepackage{amsfonts}
\usepackage{nicefrac}
\usepackage{microtype}
\usepackage{amsmath}
\usepackage{cleveref}
\usepackage{graphicx}
\usepackage{subcaption}

\title{An Evolutionary Approach for Designing Stable and Highly Expressible Low-Immunogenicity Therapeutic mRNA Sequences}

\author{
\href{https://orcid.org/0000-0003-0746-2315}
{\includegraphics[scale=0.06]{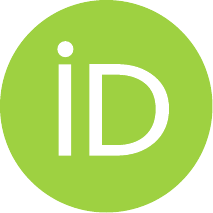}\hspace{1mm}Dhawa Sang Dong}
\\
Department of Artificial Intelligence\\
School of Engineering, Kathmandu University \\
Dhulikhel, Nepal\\
\texttt{dhawadong@gmail.com}
\and 
\href{https://orcid.org/0009-0003-7260-1702}
{\includegraphics[scale=0.06]{orcid.pdf}\hspace{1mm}Mausam Gurung}
\\
Department of Electronics and Computer Engineering\\
Kathmandu Engineering College, Trivubhan University\\
Kathmandu, Nepal\\
\texttt{mausaam.gurung593@gmail.com}
\and
\href{https://orcid.org/0000-0001-6874-9155}
{\includegraphics[scale=0.06]{orcid.pdf}\hspace{1mm}Suraj Kandel}
\\
Hetauda School of Management and Social Science\\
Trivubhan University \\
Hetauda, Nepal\\
\texttt{srjkadel@gmail.com}
}

\renewcommand{\shorttitle}{}

\hypersetup{
    pdftitle={An Evolutionary Approach for Designing Stable and Highly Translatable Low-Immunogenicity Therapeutic mRNA Sequences},
    pdfsubject={q-bio.NC, q-bio.QM},
    pdfauthor={Dhawa S. Dong, Elias D. Striatum},
    pdfkeywords={mRNA design, genetic algorithm, codon optimization, RNA stability, therapeutic mRNA}
}

\begin{document}
\maketitle

\begin{abstract}
	Messenger RNA (mRNA) sequences as therapeutics require optimized design to ensure efficient translation, structural stability, and minimal immunogenicity. This study presents a two-stage in-silico framework that integrates deep learning and evolutionary computation for rational mRNA optimization instead of existing state-of-the-art models. In the first stage, a pretrained CodonTransformer (BERT-like Large Language Model) generates biologically coherent mRNA sequences encoding the target antigen. In the second stage, a genetic algorithm (GA) evolves these candidate sequences through codon-aware crossover and synonymous mutation guided by human codon usage preferences.
    Fitness functions for evaluation combined translation-related metrics (CAI, tAI, codon-pair bias), mRNA structural stability (local and global MFE via RNAfold, GC content), and reduced immunogenicity (CpG/UpA motif frequency). Over successive generations (38th, 40th, and 42nd), the GA improved (achieved CAI values of 0.73 to 0.74 and tAI values of 0.63 to 0.64) CAI and tAI by over 6\% and codon-pair bias is high and consistent ($\sim$ 0.97) and improved ribosomal accessibility at the $5'$ end, with an unpaired\_30 fraction reaching 0.87; Global Minimum Free Energy (MFE) converged to a balanced range of -346 to -356 kcal/mol, achieving approximately 84\% base-paired structural stability, and reduced immune-stimulatory motifs - lowering the average immune penalty to 27.3 in the final generation. 
    Linear Design produces hyper-stable transcripts (MFE < -2000 kcal/mol) that risk translation inefficiency due to extreme rigidity, and BiLSTM-CRF focuses solely on high CAI (0.96 to 0.98) without structural constraints, our framework achieves an optimal translation–stability equilibrium, highlighting the proposed BERT–GA framework as an effective, data-driven approach for the design and optimization of in-silico mRNA sequences.

\end{abstract}

\keywords{mRNA design, CodonTransformer, Large Language Model, BERT, genetic algorithm, codon optimization, RNAfold, translation efficiency, immunogenicity, in-silico optimization, therapeutics, BERT-Embeddings.}

\section{Introduction}
\textbf{mRNA Sequences} 

Significant attention for mRNA-based therapeutics has been gained in treatment of infectious diseases and cancer; however, challenges such as stability, efficient translation and delivery, and immunogenicity have limited their widespread clinical applications. As studied in \cite{Ref@14, Ref@15, Ref@17}, advances in mRNA modification, optimized 5’ UTR, codon usage, lipid nanoparticle (LNP) formulations, and targeted delivery strategies have enhanced stability, translation efficiency, and so the therapeutic potential by optimizing mRNA design and delivery for next generation drug development; mRNA therapeutics is emerging as a transformative platform in modern medicine enabling rapid development of vaccine, protein replacement therapies, and cellular reprogramming strategies \cite{Ref@63}. The structural components of mRNA – including the 5’ cap Untranslated Region (UTRs), coding sequence (CDS), and poly (A) tails – are critical determinants of mRNA stability and translation efficiency, prompting ongoing efforts to optimize these regions for clinical applications \cite{Ref@63}. 

\textbf{LLM for mRNA}

The advent of Artificial Intelligence (AI), particularly Large Language Models (LLMs), is revolutionizing the landscape of drug discovery and development \cite{Ref@101, Ref@19}. These models excel at generating novel molecules, predicting pharmacological properties and optimizing synthesis pathways, thereby speeding up traditionally lengthy processes of drug design. Their ability to incorporate extensive biological, chemical and clinical data – especially when combined with external knowledge resources – enables the design of innovative therapeutic agents with greater precision and efficiency. In messenger RNA (mRNA) based drug design \cite{Ref@12}, LLMs enhance protein structure prediction, molecular generation, and precision in drug discovery and mRNA therapeutics; additionally, LLMs \cite{Ref@51} accelerate the design of optimal mRNA sequences for enhanced stability and expression. In this conetxt, \cite{Ref@52} the innovative mRNA vaccine technology in recent years has been in the focus of engineering the mRNA sequence, development of methods that enable simple, rapid and large-scale current Good Manufacturing Practice (cGMP) of mRNA, and development of highly efficient and safe mRNA vaccine delivery materials.  

In the context of mRNA-based therapeutics, AI techniques including LLMs offer immense potential: drug target identification, protein structure understanding allowing significant reduction in time and cost of drug design \cite{Ref@30}. Designing effective mRNA-based drugs involves optimizing sequences for stability, translational efficiency, and minimal immunogenicity, alongside formulating suitable delivery systems. AI-driven sequence design can analyze vast datasets to generate optimized mRNA sequences swiftly, facilitating rapid development – an advantage critically needed during pandemics and personalized medicine applications \cite{Ref@75}. Moreover, AI can assist in formulating lipid nanoparticles or alternative delivery vectors, predicting their interactions with biological systems.  (\textbf{editing done})

Recent advancements in computational approaches, notably LLMs and specialized models like UTRBERT and mRNA-LM, facilitate in-depth analysis of full-length mRNA sequences, capturing complex biological interactions and evolutionary patterns across species \cite{Ref@38, Ref@39}. These models not only improve the understanding of sequence-function relationships but also accelerate the rational design of mRNA constructs with enhanced efficacy and safety profiles, addressing some of the key challenges in large-scale manufacturing, immunogenicity, delivery, and stability \cite{Ref@63, Ref@38, Ref@13}. Additionally, LLMs \cite{Ref@59} have the ability to anticipate genomic elements such as enhancing transcription factor binding sites, and promoter regions thereby improving the precision and effectiveness of disease biomarker identification and prediction of therapeutic targets. 

\section{Literature Review}
\textbf{Secondary structure:} 

Recent work \cite{Ref@1} shows that LLMs for RNA secondary structure prediction have demonstrated the potential in sequence-based structure analysis with models like ERNIE-RNA and RiNALM with the highest accuracy. Authors introduced \cite{Ref@11} MXfold2 – deep learning algorithm – which integrates thermodynamic information with neural-folding scores improving RNA structure prediction robustness and accuracy – key for designing stable mRNA therapeutics – by training with thermodynamic regularization for broad sequence stability. 

\textbf{mRNA Optimization:}

Authors \cite{Ref@2} proposed new optimization algorithm – Linear Design which optimized mRNA sequences improving stability and codon efficiency which demonstrated significant enhancement of vaccine immunogenicity. 
Authors \cite{Ref@81} proposed a deep learning framework utilizing a BiLSTM-CRF architecture and the novel concept of "codon boxes" to autonomously learn host-specific codon distributions and significantly enhance recombinant protein expression while maintaining 100\% sequence fidelity. The mRNAid platform \cite{Ref@82} leverages the DNA Chisel framework to perform multi-objective optimization, allowing researchers to balance "hard" biological constraints like uridine depletion (no U in the third codon position) with "soft" objectives such as Minimum Free Energy (MFE) and the Codon Adaptation Index (CAI). 

LLMs like Geneformer plays a crucial role in mRNA-based drug design by analyzing large-scale transcriptomic data to map gene regulatory networks. The models improve therapeutic target identification and optimize mRNA sequence design, outperforming traditional deep learning approaches \cite{Ref@4}. Similarly, large language models like codonBERT \cite{Ref@31} enable biologically informed mRNA sequence optimization by capturing codon context, RNA structure, and evolutionary constraints, improving protein translation efficiency and stability for therapeutic applications. The study \cite{Ref@38} introduced mRNA-LM, a full-length small language model (sLM), designed for mRNA sequence optimization in drug and vaccine development. The model integrates BERT based sub models – 5UTR-BERT, codonBERT, and 3UTR-BERT to capture interactions across $5’$ UTR, CDS, and $3’$ UTR regions. Using contrastive learning (CLIP), mRNA-LM outperforms existing models (RNA-FM, Saluk) in mRNA half-life prediction, translation efficiency and protein expression. The model also shows zero-shot generalization on mRNA vaccine datasets, making it a promising tool for mRNA drug design and development. (future work – integrating secondary structure prediction and improving tokenization for longer sequences). The stability of a given mRNA Open Reading Frame (ORF) sequence \cite{Ref@53} is related to functional cognate tRNA availability which is in turn influenced by upstream amino acid concentrations. Stable mRNAs contain high proportions of codon with relatively high functional tRNA availability and are translated quickly. 

Authors in \cite{Ref@75} introduced RNop framework – a transformer -based deep learning model designed for multi-objective mRNA optimization. RNop employs four novel loss functions – GPLoss, CAILoss, tAILoss and MFELoss – that enable explicit control of sequence fidelity, codon adaptation, secondary structure stability, and translation efficiency. They achieved high computational throughput up to 47.32 sequences per second and validated through the in vivo experiment showing up to 4.6 fold increase in protein expression over original sequences demonstrating RNop’s capability for scalable, accurate mRNA design. Additionally, recent advancement in deep learning such as codon BERT architecture \cite{Ref@76}, utilize a cross-attention mechanism to effectively model the contextual relationship between codon and amino acids by training on high expression transcript data. This approach enables the model to learn and optimize codon sequences tailored to specific expression goals, achieving near-perfect accuracy and high correlation with native codon usage patterns thereby enhancing mRNA stability and expression – critical factors in mRNA therapeutic design and development. Moreover, in \cite{Ref@37} authors introduced helix-mRNA model that leverages a hybrid attention and state-space architecture with a two stage pretraining strategy on diverse datasets, enabling efficient analysis of full-length mRNA sequences to predict properties like stability and translation efficiency – outperforming existing models with fewer parameters and longer sequence processing capabilities. Authors \cite{Ref@207}[Ref\#207] developed RiboDecode model that learns ribo-seq (ribosome profiling data) to capture complex biological patters and optimize translation efficiency and therapeutic efficacy; however, the proposed model excluded UTRs while optimizing that limits precision of synthetic therapeutic design. On the other hand, authors \cite{Ref@210} [Ref\#210] introduced iDRO – first framework for full length mRNA sequences design; study implemented BiLSTM-CRF for codon selection for higher expression and RNA-BERT to generate UTR suitable for human cells; however, study is lagged to explored linear-complexity architecture to better optimization of long sequences. Moreover, Xiong et al. \cite{Ref@208} [Ref\#208] developed mRNA-BERT model using Linear Bias attention mechanism to predict protein expression, ribosome load, and stability; while training model, authors uses dual tokenization (nucleotide for UTRs and codon for CDS) and cross model contrastive learning to identify RBP binding site and m6A modification site; however, model still fails to capture higher order structural information. Addition to that, Zhang et al. \cite{Ref@72} introduced mRNA2vec, a self-supervised language model that jointly models the $5'$ UTR and CDS regions using data2vec-inspired training, incorporating auxiliary tasks like MFE and secondary structure prediction, leading to significant improvement in translation efficiency, stability and expression level predictions for mRNA therapeutics. 

\section{Methodology}
\subsection*{I. mRNA Design Framework}
We developed a two-stage in-silico drug design pipeline that integrates transformer-based sequence generation and genetic evolutionary optimization; the input sequences to the complete design framework (synthesis of mRNA sequences) encodes a given target antigen protein. The figure \ref{fig:WorkFlowDiagram} gives the complete workflow diagram for mRNA design and optimization framework – from target protein input to getting optimized mRNA sequences. 

\subsection*{a. BERT-Based Generation:} 
A pretrained transformer language model (CodonTransformer - like BERT) was used to generate initial candidate mRNA sequences which encodes the amino acid sequence for given antigen protein. The model treats codons (combination of three nucleotides) as contextual “tokens,” thereby producing biologically coherent mRNA outputs preserving amino acid identity. Each generated sequence begins with start codon AUG, and some invalid or incomplete codon sequences are discarded to avoid their further evolution. Initially, candidate sequence generation using CodonTransformer ensures the start of genetic evolutionary optimization with linguistically and biologically realistic mRNA candidates that resemble naturally expressed transcripts, but it does not consider the stability of secondary structures of mRNA explicitly, and GA optimization considers stability in the case of our optimization framework. 

\subsection*{b. Genetic Algorithm Optimization:}
The sequences generated by the CodonTransformer are subsequently evolved through a genetic algorithm (GA) which performs multi-objective optimization. The GA integrates codon-level crossover, synonymous mutation, and tournament-based selection to iteratively evolve each candidate sequences. The optimization considers multiple biological and computational metrics which include codon usage efficiency (CAI, tAI), codon-pair bias, RNA structural stability (both local and global MFE), GC content, and innate immune motif density (CpG/UpA) as given by the equation \ref{eq:fitness} below where $\tilde{f}_i(x)\in[0,1]$ gives normalized $i^{th}$ optimization criteria such as codon adaptation, translation efficiency, immunogenicity etc.; so the optimization considered full mRNA sequence as candidate sequence and fitness score has optimization function of higher translation efficiency, moderate folding structure stability and Low immunogenicity. A BERT-embedding similarity metric, BERT-Score, serves as a “naturalness” constraint, maintaining semantic proximity to biologically plausible mRNA sequence space. 
\begin{equation}
F(x)=\sum_{i=1}^{n} w_i\,\tilde{f}_i(x),
\qquad  \text{ where}  ~~~~ 
\sum_{i=1}^{n} w_i = 1
\label{eq:fitness}
\end{equation}

\begin{figure}[t]
    \centering
    \includegraphics[width=0.5\linewidth]{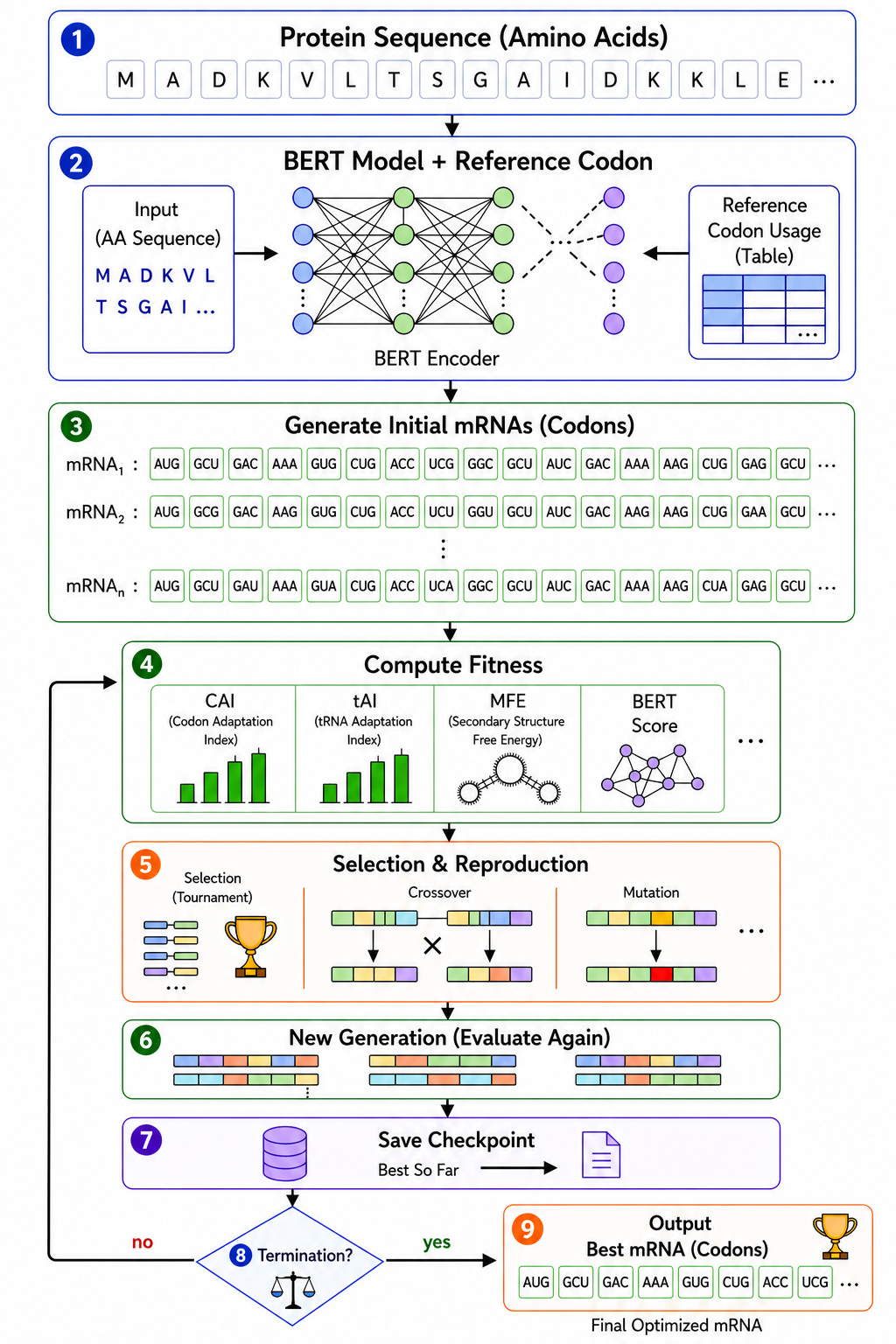}
    \caption{Work Flow Diagram}
    \label{fig:WorkFlowDiagram}
\end{figure}

Hence, the mRNA sequence design framework combines semantic intelligence on codon sequences using BERT model with genetic evolution and ensures that optimized sequences remain biologically valid using BERT embedding similarity score while achieving maximal translational and structural fitness. (\textbf{editing done})

\subsection*{II. Genetic Algorithm Design}
Candidate sequences input to the genetic algorithm correspond to coding DNA sequences (CDS) encoding the target antigen; those candidates are initialized by converting BERT-generated mRNA sequences into their corresponding DNA sequences. The GA begins with an initial population of 220 candidate sequences. To explore the high-dimensional sequence space effectively, the population size is designed to grow with each successive generation, reaching a maximum cap of 1,000 individuals. Synonymous codon substitution drives mutation, guided by human codon usage frequencies and codon weight dictionaries to preserve amino acid sequence unchanged while promoting host-preferred codons. Genetic diversity is maintained through the following hyper-parameters: mutation\_rate = 0.2; crossover\_rate = 0.8; tournament\_size = 3. The overall workflow is illustrated in Figure \ref{fig:WorkFlowDiagram}. 

\subsection*{III. Evaluation Metrics}
Each candidate mRNA construct was evaluated across weighted biological and computational parameters reflecting translation efficiency, structural stability, and immunological compatibility.

\textbf{a. Codon Usage Efficiency/translation efficiency:} Codon Adaptation Index (CAI) and tRNA Adaptation Index (tAI) were computed relative to human reference codon usage. 

CAI measures the extent to which a given sequence utilizes the most preferred codons for each amino acid, based on codon usage patterns of highly expressed human genes – higher CAI often correlates with higher protein expression. 

tAI measures how compatible a gene’s codons are with the host’s tRNA pool — codons with abundant tRNAs increase the score.

Higher CAI and tAI values indicate improved codon optimization and better adaptation to the host tRNA pool, enhancing translation efficiency.

\textbf{b. Codon-Pair Bias:} Codon-pair frequencies were analyzed to favor natural pair distributions observed in human coding sequences, supporting ribosomal efficiency and smooth translation kinetics.

\textbf{c. Translation Initiation Accessibility:} RNAfold was used to evaluate the minimum free energy (MFE) structure within ±30 nucleotides of the AUG. The accessibility of the start codon and fraction of unpaired bases in the local window were quantified.

\textbf{d. Global mRNA Structural Stability:} The minimum free energy (MFE) of each full-length mRNA sequence ($5'$ UTR + CDS + $3'$ UTR) was computed using ViennaRNA RNAfold to evaluate global folding stability (thermodynamically optimal secondary structure). Moderately negative MFE values ($\approx$ - 330 to -360 kcal/mol) were considered optimal, indicating balanced structural integrity without excessive secondary folding.  

\textbf{e. GC Content:} Global GC (number of G plus number of C) percentage of the CDS was constrained to remain within biologically plausible ranges to ensure structural stability without excessive secondary folding.

\textbf{f. Immune score:} CpG (C–Phosphate–G) and UpA (U–Phosphate–A) motif frequencies were quantified to evaluate potential innate immune activation; lower immune scores were considered favorable.

\textbf{g. UTR Balance:} The $5'$ and $3'$ UTRs were optimized to maintain balanced context ($\sim$ 0.6 – 0.7) for stable transcript structure and regulatory control.

\textbf{h. Motif Complexity:} The total number of secondary structure motifs was maintained at moderate levels ($\approx$100 ± 10) to preserve manageable folding complexity and transcription efficiency.

\textbf{i. Embedding Similarity:} Candidate sequences were embedded using a pretrained cdsBERT model. Cosine similarity to the reference sequence embedding was used to encourage naturalness.

\section{Results}
\subsection*{Generational Optimization of mRNA Constructs}
\begin{figure}[b]
    \centering
    
    \begin{subfigure}{0.48\linewidth}
        \centering
        \includegraphics[width=\linewidth]{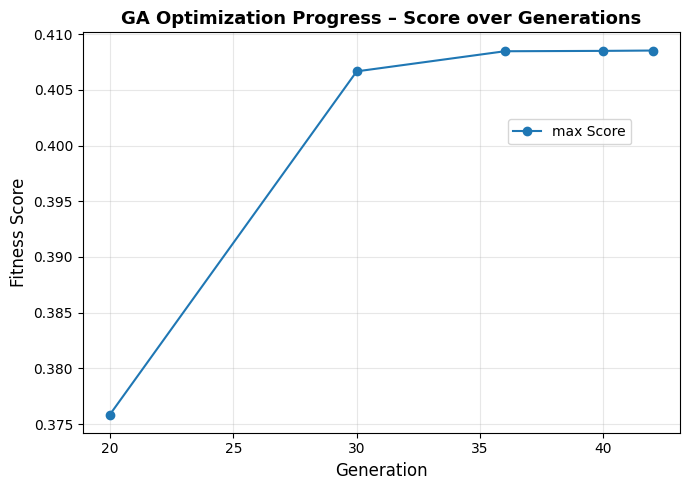}
        \caption{Evolution of Fitness Score}
        \label{fig:2a}
    \end{subfigure}
    \hfill
    \begin{subfigure}{0.48\linewidth}
        \centering
        \includegraphics[width=\linewidth]{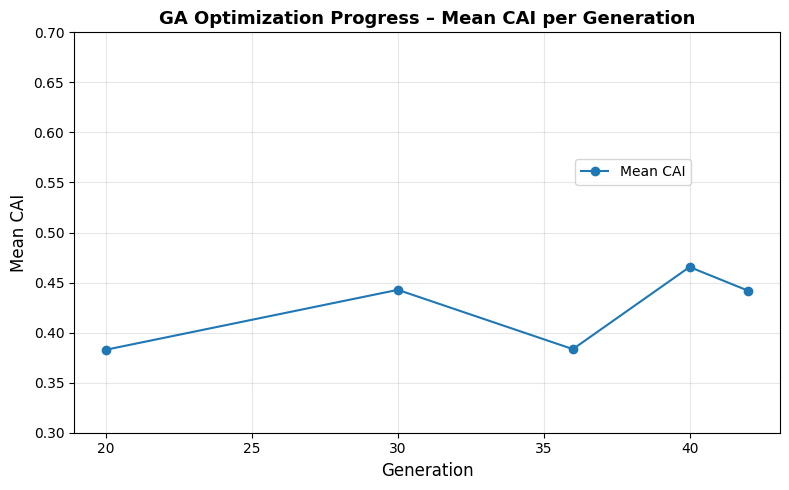}
        \caption{Evolution of Mean CAI}
        \label{fig:2b}
    \end{subfigure}
    
    \caption{GA Optimization over the Generation}
    \label{fig:GAOptimization}
\end{figure}

BERT generated mRNA sequences were evolved using evolutionary process – genetic algorithm; The formal stopping criterion for the algorithm was the saturation of the fitness score. As observed in Figure \ref{fig:2a}, the fitness score values improved significantly from the 20th generation but began to plateau as they approached the 42nd  generation. Among the evolved generations, three generations of codon-optimized mRNA sequences ($38^{th}$, $40^{th}$, and $42^{nd}$) were analyzed using a genetic algorithm (GA) pipeline subjected to enhance translation efficiency (CAI, tAI, Codon-pair Bias and AUG Accessibility), structural stability (Local MFE, Global MFE and GC content), and immune tolerance (Immune motif score). Each generation represents a population of top-ranked sequences filtered by composite fitness functions integrating CAI, tAI, codon-pair score, global MFE, immune score, GC balance, and UTR accessibility.

\begin{table}[t]
\centering
\caption{Evolution of optimization metrics across generations}
\label{tab:evolution_metrics}

\resizebox{\textwidth}{!}{%
\begin{tabular}{lcccc}
\toprule
\textbf{Parameter} & \textbf{38th Generation} & \textbf{$40^{th}$ Generation} & \textbf{$42^{nd}$ Generation} & \textbf{Evolutionary Trend} \\
 & \textbf{(Mean $\pm$ SD)} & \textbf{(Mean $\pm$ SD)} & \textbf{(Mean $\pm$ SD)} & \\
\midrule
CAI & $0.692 \pm 0.010$ & $0.704 \pm 0.008$ & $0.738 \pm 0.009$ & $+6.6\%$ \\

tAI & $0.611 \pm 0.015$ & $0.625 \pm 0.012$ & $0.635 \pm 0.010$ & Improved adaptation \\

Global MFE (kcal/mol) & $-372 \pm 18$ & $-362 \pm 16$ & $-354 \pm 14$ & Converging to ideal range ($-330$ to $-360$) \\

GC Content & $0.685 \pm 0.012$ & $0.671 \pm 0.011$ & $0.660 \pm 0.010$ & Stabilized near optimal ($\sim 0.66$) \\

Immune Penalty & $31.2 \pm 3.1$ & $29.4 \pm 2.8$ & $27.3 \pm 2.5$ & Reduced immune activation \\

Immunogenic Motifs & $119 \pm 8$ & $112 \pm 7$ & $106 \pm 7$ & Decreased motif count \\

Unpaired\_30 Fraction & $0.70 \pm 0.05$ & $0.81 \pm 0.04$ & $0.87 \pm 0.03$ & Improved ribosomal accessibility \\
\bottomrule
\end{tabular}%
}
\end{table}

From the Figure \ref{fig:GAOptimization}, it was observed that genetic optimization improved the fitness score (figure \ref{fig:GAOptimization}a) values while evolving from generation $20^{th}$ to $42^{nd}$  generation; and mean CAI score is improving while evolving as demonstrated in \ref{fig:GAOptimization}b. Generation 42 was specifically selected as the terminal point because the composite fitness had effectively converged, and further evolution yielded negligible improvements in parameters like CAI and Global MFE. 

As shown in Table \ref{tab:evolution_metrics}, a noticeable evolutionary improvement was witnessed across generations under study: the mean CAI increased from 0.692 ± 0.01 (38th gen) to 0.704 ± 0.008 (40th gen), reaching 0.738 ± 0.009 (42nd gen) - representing a 6.6\% improvement in codon adaptation in relative to the 38th generation while evolving to 42nd generation. Similarly, tAI improved from 0.611 to 0.635, indicating better adaptation to host tRNA abundance. Codon-pair bias remained consistently high across all generations and was therefore excluded from Table \ref{tab:evolution_metrics} to focus on metrics showing meaningful variation.  

Regarding to structural stability of candidate mRNA sequences evolved, global MFE values converged to the biologically preferred range (-330 to -360 kcal/mol), with the 42nd generation demonstrating the tightest clustering (-354 ± 14 kcal/mol). The GC fraction also converged near an optimal 0.66 ± 0.01, suggesting that the GA successfully moderated excessive GC enrichment while maintaining folding integrity.  
Immune-related penalties showed a modest reduction, with average immune scores decreasing from 31.2 ($38^{th}$ generation) to 27.3 ($42^{nd}$ gen), and immunogenic motifs declining from 119 ± 8 to 106 ± 7. This shift was accompanied by an increase in unpaired\_30 fractions (0.70 to 0.87), improving ribosomal accessibility near the $5'$ end.

\subsection*{Selection of Optimal Constructs for Wet-Lab Testing}

Based on the composite evaluation model, each sequence was scored against the ideal ranges defined for high translational efficiency and moderate folding stability. Within the 40th generational evolution, Seq 4, Seq 1, and Seq 6 demonstrated the best parameter balance (CAI $\approx$ 0.70, MFE $\approx$ -340 kcal/mol, immune $\approx$  12 -- 13) and could be shortlisted as early wet-lab mRNA sequence candidates. 

The subsequent evolutionary refinement in the 42nd generation produced higher-performing constructs with improved translational efficiency and more accessible coding sequences, as lower MFE values indicated greater structural stability and enhanced ribosomal accessibility.  Among these, Seq\_ID 2, 6, and 9 consistently ranked highest in composite fitness. 

\begin{figure}[b]
    \centering
    \includegraphics[width=0.5\linewidth]{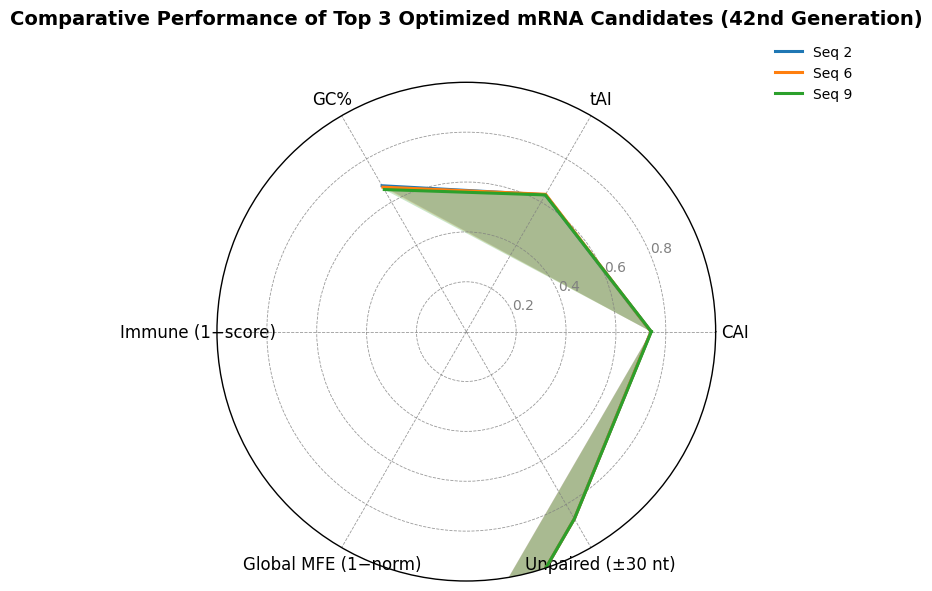}
    \caption{Comparative radar plot of the top three optimized mRNA candidates (seq 2, 6, and 9) from the $42^{th}$ generation}
    \label{fig:RadarPlot}
\end{figure}

These three sequences collectively demonstrated the best trade-off between translational efficiency as in figure \ref{fig:RadarPlot}; and immune safety, making them the most suitable candidates for downstream wet-lab experiments; including in-vitro transcription and expression studies. 

\begin{table}[t]
\centering
\caption{Comparison of selected optimized sequences}
\label{tab:ParamReadings}

\resizebox{\textwidth}{!}{%
\begin{tabular}{lccccccp{5cm}}
\toprule
\textbf{Sequence} & \textbf{CAI} & \textbf{tAI} & \textbf{MFE (kcal/mol)} & \textbf{Immune} & \textbf{GC} & \textbf{Unpaired\_30} & \textbf{Notes} \\
\midrule
Seq 2 & 0.7406 & 0.6340 & $-356.2$ & 26.18 & 0.676 & 0.8667 & Balanced translation and stability \\

Seq 6 & 0.7408 & 0.6360 & $-346.4$ & 28.94 & 0.669 & 0.8667 & Best tAI; strong stability \\

Seq 9 & 0.7419 & 0.6333 & $-354.5$ & 28.57 & 0.658 & 0.8667 & Highest CAI; optimal GC \\
\bottomrule
\end{tabular}%
}
\end{table}

As presented in figure \ref{fig:RadarPlot}, each axis represents normalized biological metrics including translation efficiency (CAI, tAI), GC content, immune score, global folding stability (MFE), and start-codon accessibility (Unpaired ±30 nt). All candidates show closely aligned high-performance profiles, with Seq 9 excelling in codon adaptation, Seq 6 in tRNA compatibility and structural stability, and Seq 2 achieving a balanced trade-off across all. 

Figure \ref{fig:SecStructRNA} shows the predicted mRNA secondary structures for the three optimized sequences (Seq 2, 6, 9), highlighting moderately stable global folding and open regions near the start codon. These predicted folds exhibit balanced stem-loop formations with unpaired regions around the $5'$ end, consistent with the improved unpaired\_30 fraction observed in Table \ref{tab:ParamReadings}. Such structures suggest enhanced ribosomal accessibility and efficient initiation, aligning with the higher CAI and tAI values recorded for the $42^{nd}$ generation constructs.

\section{Discussion and Conclusion}
The gradual improvement in codon adaptation indices (CAI) and folding energy over the last three generations ($38^{th}$, $40^{th}$, and $42^{nd}$) validates the evolutionary genetic algorithm in optimized mRNA sequence design in terms of translation efficiency, stability and immunogenicity. Incremental shifts in CAI and tAI reflect a consistent convergence toward host-preferred codons, thereby enhancing potential translation speed and accuracy. The concurrent moderation of global MFE (from highly negative to moderately negative) indicates that the algorithm avoided overly rigid secondary structures that can hinder ribosome accessibility.

\begin{figure}[b]
    \centering
    \includegraphics[width=0.5\linewidth]{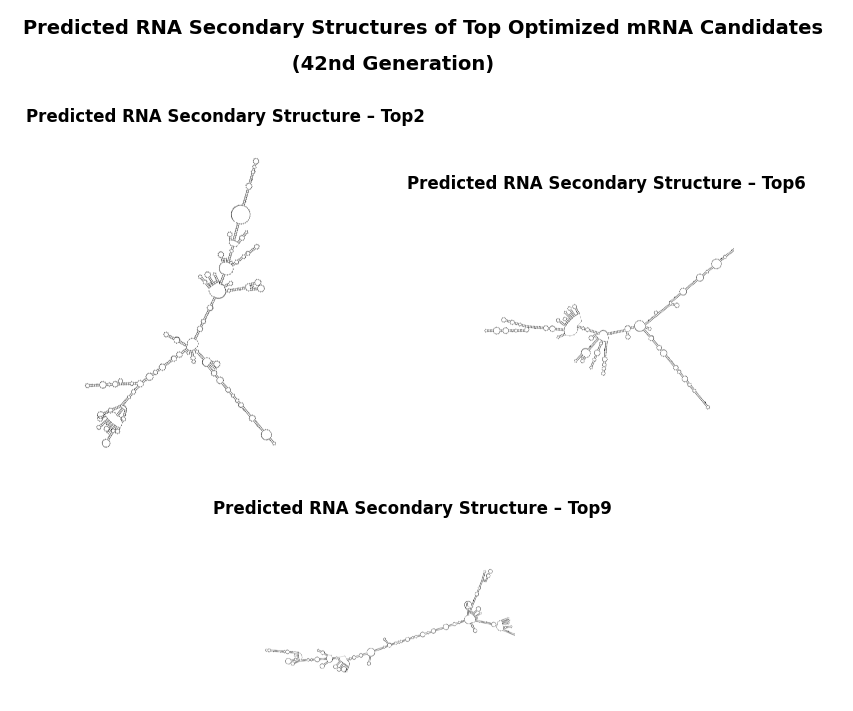}
    \caption{mRNA Secondary Structures}
    \label{fig:SecStructRNA}
\end{figure}

The predicted secondary structures of potentially the top-performing mRNA constructs (Seq 2, 6, 9), shown in Fig. \ref{fig:SecStructRNA}, further validate these numerical findings. Numerically, these three sequences exhibit balanced folding topologies with moderately stable stem-loop arrangements and open regions near the $5'$ end, consistent with the increased unpaired\_30 fractions and optimal global MFE values. These structural configuration support enhanced ribosomal accessibility, complementing the improvements in CAI and tAI observed during evolutionary optimization -- numerical findings. 

The $42^{nd}$ generation has achieved a significant balanced GC fraction ($\sim$ 0.66), consistent with the literature’s optimal window (0.55 - 0.60) for mammalian translation efficiency. Though GC values remained slightly above the theoretical maxima, the folding energy and unpaired-region accessibility suggest compensatory structural flexibility at the $5'$ end. From an immunological standpoint, while immune penalties remained within the “acceptable” range (25 – 29), the observed decline in CpG and U-rich motifs across generations implies that the GA was partially effective while reducing innate immune activation, so the moderate stability.

Hence, this trend supports that computational evolution can fine-tune mRNA design not only for expression efficiency but also for reduced immunogenicity, a critical feature for therapeutic mRNA development. The three potential candidates (Seq 2, 6, 9) exhibit complementary strengths: Seq 2 shows the best overall balance between CAI, MFE, and immune metrics; Seq 6 demonstrates superior tAI, indicating potentially faster ribosomal elongation; and Seq 9 achieves maximal codon adaptation with a well-balanced GC content. Collectively, they represent diverse design solutions within a narrow high-fitness landscape, which is advantageous for experimental validation.

Table \ref{tab:T3_model_comparison} summarizes the comparative analysis and highlighted the distinctive advantages of the proposed BERT–GA framework in multi-objective landscape of mRNA design. By integrating transformer-based biological semantic embeddings with an evolutionary genetic algorithm, our approach achieves a superior balance between translational efficiency and structural integrity with reduced immunogenicity addressing the limitations observed in existing state-of-the-art models.

\subsection*{The Stability–Efficiency Paradox}
The major challenge in mRNA engineering is the trade-off between codon optimality and Minimum Free Energy (MFE). As demonstrated in Table \ref{tab:T3_model_comparison}, the Linear Design algorithm \cite{Ref@2} identifies "hyper-stable" transcripts with extremely low MFE values (-2487.3 kcal/mol) maximizing half-life. However, analysis suggests that such hyper-stable secondary structures may introduce a risk of translation inefficiency creating overly rigid constructs against ribosome accessibility. As a solution to the problem, the BERT – GA framework targets a "balanced" MFE range of -346 to -356 kcal/mol. Despite these moderate MFE energy values, our optimized sequences (specifically the 42nd generation) achieved structural stability of approximately 84\% base-pairing which is comparable to the stability of hyper-stable constructs by Linear Design while maintaining the "structural flexibility" requirement for efficient ribosomal elongation and protein translation initiation.

\subsection*{Multi-Metric Codon Optimization}
Though models like BiLSTM-CRF \cite{Ref@81} achieve exceptionally high CAI values (0.96 - 0.98), they often lack explicit structural or immunological constraints. The BERT–GA framework demonstrated a 6.6\% improvement in CAI across generations, converging at 0.73 - 0.74. Uniquely, our optimization framework also incorporates the tRNA Adaptation Index (tAI), achieving values in the range of 0.63 - 0.64. The inclusion of tAI ensures that the sequence is optimized not just for general codon preference, but specifically for the host tRNA pool abundance, which is a critical determinant of translation efficiency and accuracy. 

\begin{table}[t]
\centering
\caption{Comparison of proposed framework with related mRNA optimization approaches}
\label{tab:T3_model_comparison}

\resizebox{\textwidth}{!}{%
\begin{tabular}{p{3.2cm}ccccp{6.5cm}}
\toprule
\textbf{Model} &
\textbf{CAI (Spike mRNA)} &
\textbf{tAI} &
\textbf{Global MFE} &
\textbf{Structural Stability} &
\textbf{Interpretation} \\
&
&
&
\textbf{(kcal$\cdot$mol$^{-1}$)} &
\textbf{(base-paired \%)} &
\\
\midrule

LinearDesign \cite{Ref@2} &
$0.74\;(\lambda = 0)\rightarrow0.668\;(\lambda=\infty)$ &
N/A &
$-2487.3 \rightarrow -2031.8$ &
$83.6\% \rightarrow 75.3\%$ &
Designed SARS-CoV-2 Spike mRNAs show a clear CAI--MFE trade-off; higher $\lambda$ yields codon-optimized yet hyper-stable transcripts, while extreme structures may reduce translation efficiency. \\

BiLSTM-CRF \cite{Ref@81} &
$0.96 \rightarrow 0.98$ &
N/A &
N/A &
N/A &
Protein expression was $11.1\times$ higher than the original for FALVAC-1. \\

CAI + U-Depletion \cite{Ref@82} &
$0.62$ &
N/A &
$-1451.70$ &
High &
Exceeds commercial vaccine performance; MFE correlates strongly with protein output. \\

Proposed BERT--GA Framework &
$0.73 \rightarrow 0.74$ &
$0.63 - 0.64$ &
$-346.40 \rightarrow -356.20$ &
$\approx 84\%$ base-paired; balanced MFE &
Integrates BERT semantic embeddings with genetic evolution and yields an optimal translation--stability balance. \\

\bottomrule
\end{tabular}%
}
\end{table}

\subsection*{Biological Realism via BERT Embeddings}
A defining feature of our framework is the use of pretrained CodonTransformer (BERT-like) embeddings serving as a "naturalness" constraint. Traditional algorithms often rely on heavy GC-enrichment, leading to biologically unrealistic codon sequences. Table \ref{tab:T3_model_comparison} illustrates that while other methods focus on extreme metric maximization, our framework yields sequences that resemble naturally expressed human transcripts -- "biologically realistic foundation" ensuring the optimized mRNA and more likely to be compatible with host cellular environment in therapeutic applications.

\subsection*{Immunological and Structural Synergy}
From the empirical values in Table \ref{tab:T3_model_comparison}, our framework performs better to minimize innate immune activation. With Sequence evolution, the frequencies of the CpG and UpA motif have been reduced (reducing an average immune penalty of 27.3 in $42^{nd}$ generation), so the proposed model effectively fine-tunes the constructs for both safety and expression. This is further evidenced by the Unpaired\_30 Fraction (0.87), which signifies high ribosomal accessibility at the $5'$ end, a feature that complements the moderate global MFE to facilitate efficient translation initiation.

Collectively, our evolutionary mRNA design framework introduces a more reliable in-silico mRNA sequence design (multi-objective optimization, reliable computational framework, high performance mRNA therapeutic and vaccines development process) pipeline than single-objective or extreme-stability models (Linear Design, BiLSTM-CRF and CAI+U-Depletion). Our BERT-GA framework of mRNA design approach achieves superior balance between translational efficiency and structural control, providing biologically more realistic foundation (closely resembled codon usage and compositional profiles of human coding DNA) for rational mRNA vaccine and therapeutic design -- achieved high base-pairing stability alongside host-optimized codon usage and reduced immunogenicity. Over the three evolutionary generations ($38^{th}$, $40^{th}$ and $42^{nd}$), the algorithm improved features CAI and tAI by higher than 6\%, optimized MFE toward functional stability (better folding), and modestly reduced immune-stimulatory motifs.
Data-driven analysis identified three top-performing constructs: Seq\_ID 2, Seq\_ID 6, and Seq\_ID 9 from the $42^{nd}$ generation of the GA-based mRNA optimization process -- as the most suitable candidates for wet-lab synthesis and expression experiments. The predicted secondary structures as shown in figure \ref{fig:SecStructRNA} reinforce these findings, showing balanced folding with accessible $5'$ regions which supports efficient translation initiation. 


\bibliographystyle{IEEEtran}
\bibliography{references}  

\end{document}